\pdfoutput=1
\documentclass[11pt]{article}

\usepackage[final]{acl}

\usepackage{times}
\usepackage{latexsym}

\usepackage[T1]{fontenc}
\usepackage[utf8]{inputenc}

\usepackage{microtype}

\usepackage{inconsolata}

\usepackage{graphicx}
\usepackage{tabularx}
\usepackage{booktabs}
\usepackage{multirow}
\usepackage{tcolorbox}
\usepackage{amsmath}
\usepackage{amssymb}

\title{Efficient Training for Cross-lingual Speech Language Models}

\author{
    Yan Zhou \textsuperscript{\rm 1,2,3},
    Qingkai Fang \textsuperscript{\rm 1,2,3},
    Yun Hong \textsuperscript{\rm 1,2,3},
    Yang Feng\textsuperscript{\rm 1,2,3}\footnotemark[2] \\
    \textsuperscript{\rm 1}{Key Laboratory of Intelligent Information Processing, Institute of Computing Technology,} \\ Chinese Academy of Sciences (ICT/CAS) \textsuperscript{\rm 2} {State Key Laboratory of AI Safety,} \\ Institute of Computing Technology, Chinese Academy of Sciences \\
    \textsuperscript{\rm 3} {University of Chinese Academy of Sciences, Beijing, China} \\
    \texttt{\href{mailto:zhouyan23z@ict.ac.cn}{zhouyan23z@ict.ac.cn}, \href{mailto:fengyang@ict.ac.cn}{fengyang@ict.ac.cn}}
}

\begin{document}
\maketitle

\renewcommand{\thefootnote}{\fnsymbol{footnote}} %将脚注符号设置为fnsymbol类型，即特殊符号表示
\footnotetext[2]{Corresponding author: Yang Feng.} %对应脚注[1]
\renewcommand{\thefootnote}{\arabic{footnote}}

\begin{abstract}
Currently, large language models (LLMs) predominantly focus on the text modality. To enable more natural human-AI interaction, speech LLMs are emerging, but building effective end-to-end speech LLMs remains challenging due to limited data and the difficulty in expanding to more languages. In this paper, we introduce \textbf{C}ross-lingual \textbf{S}peech \textbf{L}anguage \textbf{M}odel (\textbf{CSLM}), an efficient training method for cross-lingual speech LLMs based on discrete speech tokens. We propose a novel alignment strategy that achieves cross-modal and cross-lingual alignment through continual pre-training. By conducting instruction fine-tuning following a speech-text interleaved chain-of-modality generation process, we enhance modal alignment at a finer granularity, thereby improving generation quality and reducing latency. CSLM aligns different modalities and languages simultaneously without the need for massive speech data, thus exhibiting good language scalability. Evaluations on cross-modal tasks, mono-lingual conversational tasks, and cross-lingual conversational tasks demonstrate CSLM's strong cross-modal alignment capabilities and general task abilities. \footnote{\url{https://github.com/ictnlp/CSLM}}
% This document is a supplement to the general instructions for *ACL authors. It contains instructions for using the \LaTeX{} style files for ACL conferences.
% The document itself conforms to its own specifications, and is therefore an example of what your manuscript should look like.
% These instructions should be used both for papers submitted for review and for final versions of accepted papers.
\end{abstract}

\section{Introduction}
\label{sec:introduction}

In recent years, the evolution of large language models (LLMs) like ChatGPT \citep{chatgpt_introduction} has enabled the rapid development of sophisticated text-based chatbots. However, as applications of LLMs continue to expand, there is growing interest in exploring more natural human-AI interaction paradigms and unlocking the models' potential in other modalities. The emergence of speech LLMs addresses this demand, as speech-based interaction offers inherent convenience and conveys additional information beyond textual communication.

The construction of speech LLMs faces several challenges. The speech modality contains significantly more information than text modality, making speech modeling difficult. A more challenging aspect is the scarcity of speech data compared to text data, especially for certain languages. To address these issues, current researchers typically integrate text LLMs to build speech LLMs, leveraging the language capabilities and general knowledge from the text modality. The simplest approach is to cascade an automatic speech recognition (ASR) model, a text LLM, and a text-to-speech (TTS) model, but this approach brings in error accumulation and increased latency. Therefore, researchers are now focusing more on end-to-end speech LLMs. Some researchers train auto-regressive models using only speech data following the training process of text LLMS \citep{lakhotia-etal-2021-generative, NEURIPS2023_c859b99b}, but this approach suffers from the scarcity of speech data. Some researchers propose modular speech LLMs that establish mappings between existing speech encoders and text LLMs \citep{chu2023qwenaudioadvancinguniversalaudio, xie2024miniomnilanguagemodelshear, fang2024llamaomniseamlessspeechinteraction, wang2024freezeomnismartlowlatency}, but these methods fall short in speech generation and lack intrinsic speech-text alignment, limiting their general applicability. Other researchers explore unified modeling of speech and text \citep{zhang-etal-2023-speechgpt, zhan2024anygptunifiedmultimodalllm, 10.1162/tacl_a_00728, défossez2024moshispeechtextfoundationmodel, zeng2024glm4voiceintelligenthumanlikeendtoend} based on speech discretization techniques.

While exploring methods for modeling the speech modality, research on extending speech LLMs to more languages has also gained increasing attention. Many languages already face the problem of resource scarcity in the text modality, and this problem is even more severe in the speech modality. Building unified multilingual and multimodal representations typically requires massive amounts of data, thus how to achieve effective cross-lingual and cross-modal alignment simultaneously with limited data has become a core challenge.

Recognizing the critical challenges of data efficiency and cross-lingual capability in developing speech LLMs, we propose an efficient training method for cross-lingual speech LLMs. Based on a speech LLM architecture utilizing discrete speech tokens, we introduce a novel alignment strategy that achieves cross-lingual and cross-modal alignment, and conduct continual pre-training with limited data. Subsequently, we conduct instruction fine-tuning following a speech-text interleaved chain-of-modality generation process to leverage cross-modal alignment at a finer granularity, thereby improving generation quality and reducing latency. The trained general \textbf{C}ross-lingual \textbf{S}peech \textbf{L}anguage \textbf{M}odel \textbf{(CSLM)} can align different modalities and languages simultaneously without the need for massive speech data, thus exhibiting good language scalability in terms of data requirements and training difficulty. Evaluations on cross-modal tasks, mono-lingual and cross-lingual conversational tasks demonstrate CSLM's strong cross-modal alignment capabilities and general task abilities, validating the effectiveness of the proposed method.

Unlike existing models like SPIRIT LM \citep{10.1162/tacl_a_00728}, Moshi \citep{défossez2024moshispeechtextfoundationmodel} and GLM-4-Voice \citep{zeng2024glm4voiceintelligenthumanlikeendtoend} which often require extensive data, CSLM introduces an efficient method for robust cross-modal and cross-lingual alignment. Furthermore, our novel interleaved chain-of-modality fine-tuning significantly enhances generation quality and reduces latency. Our contributions are as follows:
\begin{itemize}
    \item We propose an efficient training method for cross-lingual speech LLMs that simultaneously achieves cross-lingual and cross-modal alignment without the need for huge amount of speech data.
    \item We introduce a speech-text interleaved chain-of-modality generation method for instruction fine-tuning to enhance modal alignment at a finer granularity, thereby improving generation quality and reducing latency.
    \item Our training method is easily scalable to other languages in terms of data volume and training difficulty, providing valuable guidance for training multilingual speech LLMs.
\end{itemize}

\section{Related Works}
\subsection{Speech Tokenization}
\label{sec:speech-tokenization}

Speech tokenization is the process of obtaining discrete speech tokens from continuous speech waveforms. After the discrete speech tokens are extracted, they can be used like text tokens and allow for joint modeling of speech and text.

Current speech tokenization technologies primarily employ k-means, VQ (Vector Quantization), or RVQ (Residual Vector Quantization) to obtain discrete speech tokens. \citet{hsu2021hubert} extracts semantic tokens using k-means method on self-supervised learning representations. \citet{chung2021w2v} and \citet{huang2023repcodec} obtain semantic tokens via VQ. \citet{9625818} and \citet{defossez2022high} utilize RVQ to obtain acoustic tokens, while recently SpeechTokenizer \citep{zhang2024speechtokenizerunifiedspeechtokenizer} and Moshi \citep{défossez2024moshispeechtextfoundationmodel} further use different RVQ layers to obtain both semantic and acoustic tokens. Cosyvoice \citep{du2024cosyvoicescalablemultilingualzeroshot} introduces advanced speech tokenization techniques, representing speech with supervised semantic tokens derived from a speech recognition model via vector quantization, which enables semantic decoding and high-quality speech synthesis.

\subsection{Speech LLM}
\label{sec:speech-llm}

Speech LLMs refer to LLMs that can interact with humans in speech. Depending on the modalities supported and the different approaches to modeling speech, several distinct paradigms of speech LLM have emerged, including speech-only models, modular models combining a speech encoder and an LLM, and speech-text models which jointly model discrete speech tokens and text tokens.

GSLM \citep{lakhotia-etal-2021-generative} first proposes an LLM trained solely on speech, utilizing discrete speech units to train a decoder model by predicting the next token. Similarly, TWIST \citep{NEURIPS2023_c859b99b} adopts a warm-start strategy, continuing to train a text LLM on speech data. Although these speech-only large models have the ability to model contextual relationships in speech, the sheer amount of data in the text modality naturally surpasses that in the speech modality. As a result, speech-only models can hardly achieve the same level of general task performance as text LLMs.

Qwen-Audio \citep{chu2023qwenaudioadvancinguniversalaudio} connects a pre-trained speech encoder with a pre-trained text LLM, aligning speech representations with the text LLM to achieve speech understanding. However, this paradigm is unable to accomplish speech generation. Building on this foundation, Mini-Omni \citep{xie2024miniomnilanguagemodelshear}, LLaMA-Omni \citep{fang2024llamaomniseamlessspeechinteraction} and Freeze-Omni \citep{wang2024freezeomnismartlowlatency} further add a speech synthesis model after the text LLM to generate speech. These speech LLMs, which consist of a speech encoder combined with a text LLM (some coupled with a speech synthesis model), exhibit disadvantages in terms of the quality and diversity of generated speech.

Other works have attempted to jointly model discrete speech tokens with text. SpeechGPT \citep{zhang-etal-2023-speechgpt} first proposes such a method, expanding discrete speech tokens into LLM's vocabulary. AnyGPT \citep{zhan2024anygptunifiedmultimodalllm} follows this approach and improves upon the discrete speech tokens by modeling them with separate semantic and acoustic information. Moshi \citep{défossez2024moshispeechtextfoundationmodel} proposes a full-duplex working mode of speech LLM under this paradigm. GLM-4-Voice \citep{zeng2024glm4voiceintelligenthumanlikeendtoend} first proposes a bilingual (Chinese-English) speech LLM with massive amount of data under this paradigm, and it is capable of interacting with humans through interleaved outputs of speech and text. These models require a large amount of training data and have achieved commendable results on mono-lingual speech tasks, but their cross-lingual alignment abilities are still limited. Therefore, it is difficult for them to perform some cross-lingual speech tasks.

\section{Model: CSLM}
\begin{figure}[t]
    \centering
    \includegraphics[width=\linewidth]{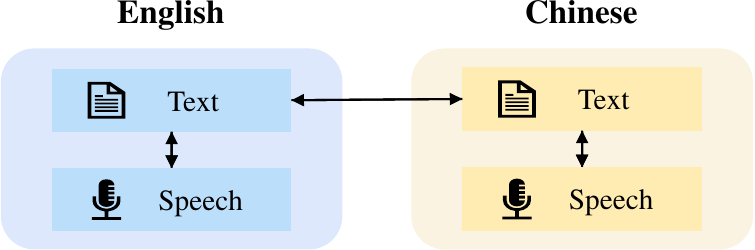}
    \caption{Alignment strategy of CSLM.}
    \label{fig:alignment}
\end{figure}
\begin{figure*}[t]
    \centering
    \includegraphics[width=\linewidth]{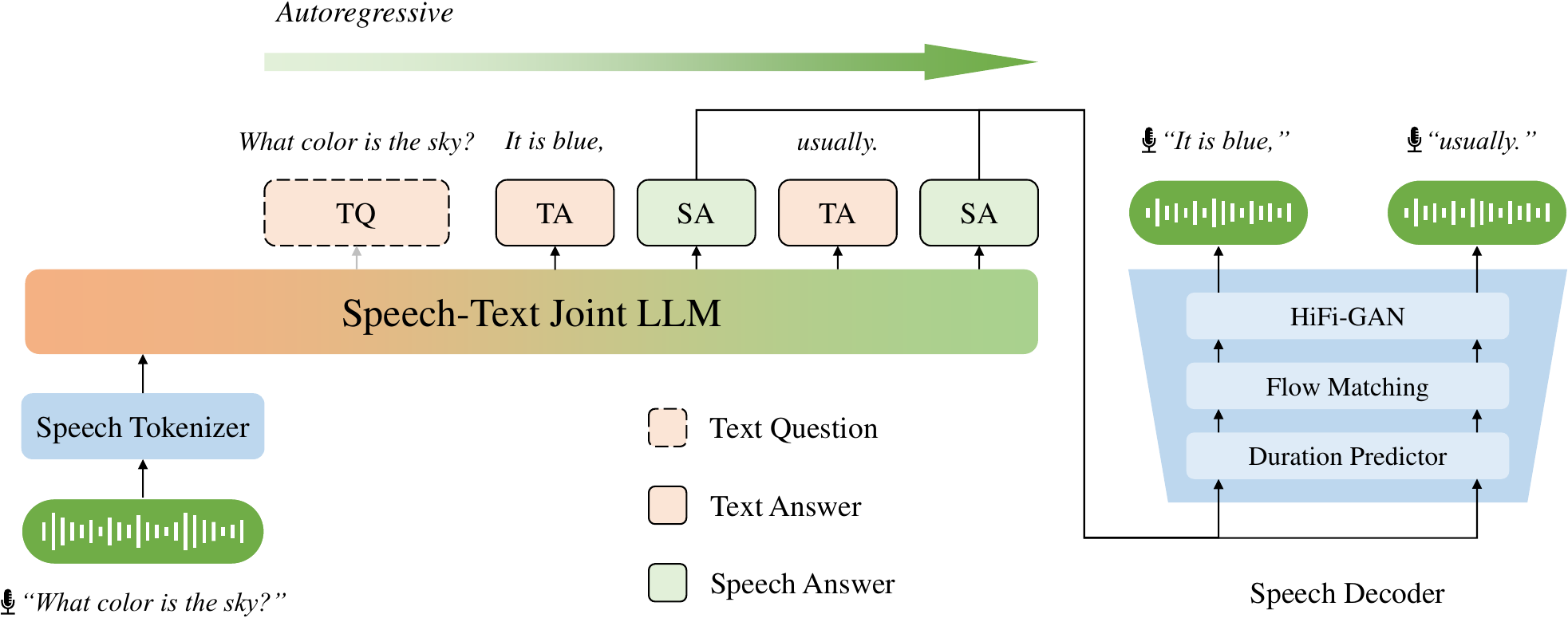}
    \caption{Model architecture and inference process of CSLM.}
    \label{fig:architecture}
\end{figure*}

CSLM is a speech LLM based on discrete speech tokens, designed to achieve both cross-modal and cross-lingual alignment. We will introduce the architecture of CSLM, the training procedure, and the inference time workflow of this model. In addition, we will elaborate on the CSLM's possibility to be extended to more languages.

\subsection{Model Architecture}
\label{sec:model-architecture}

CSLM consists of a speech tokenizer, an LLM, and a speech decoder. The speech tokenizer first extracts a speech waveform into discrete speech tokens, which are then modeled by the LLM to generate new speech tokens. Finally, these tokens are synthesized into a new waveform by the speech decoder.

\begin{itemize}
    \item \textbf{Speech Tokenizer} We use the speech tokenizer of CosyVoice-300M-25hz \citep{du2024cosyvoicescalablemultilingualzeroshot}, which has a speech vocabulary of 4096 tokens, with a frequency of 25Hz. This speech tokenizer includes a Conformer \citep{gulati20_interspeech} encoder and a vector quantization module, which can transform the input speech Mel-spectrogram into discrete vectors. For training and generation efficiency, consecutive repeated speech tokens are merged before these tokens are fed into the LLM. Note that this merging operation does not introduce semantic loss, as it primarily operates on the acoustic domain.
    \item \textbf{Speech-Text Joint LLM} Following \citet{zhang-etal-2023-speechgpt}, we merge the vocabulary from the speech tokenizer with the vocabulary of a text LLM. This integration enables joint modeling of text and speech within the LLM.
    \item \textbf{Speech Decoder} The speech decoder consists of a conditional flow matching model and a HiFi-GAN \citep{NEURIPS2020_c5d73680} vocoder from the CosyVoice decoder, with an additional convolutional module called the duration predictor. For a reduced sequence of speech tokens, the duration predictor predicts whether each token should be repeated a specified number of times and outputs the expanded speech token sequence. This sequence is then input into the flow matching model to generate the Mel-spectrogram, which is later utilized by the HiFi-GAN vocoder to synthesize the final speech waveform.
    
\end{itemize}

\subsection{Alignment Strategy}
\label{sec:alignment-strategy}

The goal of CSLM is to simultaneously achieve cross-modal alignment and cross-lingual alignment. The alignment strategy we designed is illustrated in the Figure \ref{fig:alignment}. Within a single language, cross-modal alignment is performed between speech and text, while across different languages, alignment is achieved through the text modality.

\subsection{Training Procedure}
\label{sec:training-procedure}

We adopt a two-stage training paradigm, which includes the continual pre-training stage and the supervised fine-tuning stage.

\subsubsection{Continual Pre-training}
At this stage, we begin with an LLM already fine-tuned on instruction data, and merge the speech vocabulary with its own vocabulary. We collect parallel speech-text data in different languages to achieve speech-text cross-modal alignment. Some of the data take speech as input and text as output, corresponding to the ASR task, while the rest of the data take text as input and speech as output, corresponding to the TTS task. We collect machine translation (MT) data between Chinese and English to facilitate cross-lingual alignment. Additionally, we also collect mono-lingual instruction data in both Chinese and English, so as to reduce performance degradation in the text modality. We train the model on the aforementioned data to obtain the CSLM-base model.

\subsubsection{Supervised Fine-tuning}
In the second stage, the instruction fine-tuning stage, we train the CSLM-base model on text instruction and speech-to-speech conversational data, resulting in the CSLM-SFT model.

To further align speech and text and achieve higher generation efficiency, we propose a speech-text interleaved chain-of-modality based on the chain-of-modality from \citet{zhang-etal-2023-speechgpt}. In the original chain-of-modality, after the model receives a speech question, it generates a response in the order of text question, text answer, and speech answer, i.e., $\text{TQ} \rightarrow \text{full TA} \rightarrow \text{full SA}$. In our improved speech-text interleaved chain-of-modality, the model generates only a shorter chunk of the text answer, then immediately generates the corresponding speech answer, and this cycle repeats until the end, i.e., $\text{TQ} \rightarrow \text{TA} \rightarrow \text{SA} \rightarrow \text{TA} \rightarrow \text{SA} \cdots$.

\begin{figure}[t]
    \centering
    \includegraphics[width=\linewidth]{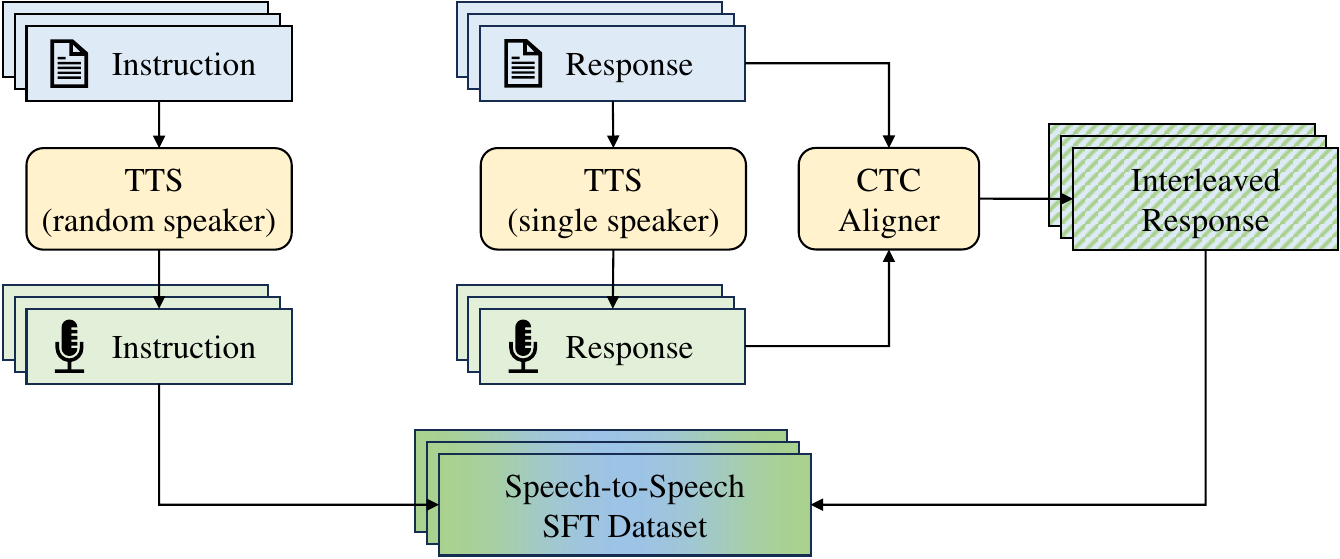}
    \caption{The construction process of the SFT dataset.}
    \label{fig:sft_construction}
\end{figure}

To construct such interleaved speech-text data, we first synthesize the instructions and responses from existing textual instruction datasets into speech, and then use a Connectionist Temporal Classification (CTC) \citep{10.1145/1143844.1143891} aligner module to obtain interleaved responses. Specifically, for a given speech-text pair $(X, Y)$, we first use the speech encoder of an ASR model to obtain the representation of the speech. Let $X = (x_1,...,x_T)$ denote raw speech inputs, through speech encoder $f_\theta$ we obtain:
\begin{equation}
    H = f_\theta(X) = (\mathbf{h}_1,...,\mathbf{h}_T) \in \mathbb{R}^{T \times d}
\end{equation}
where $d$ is the dimension of $f_\theta$. Text labels are tokenized as $Y = (y_1,...,y_L)$ with $L \ll T$. We apply CTC dynamic programming algorithm to establish the optimal alignment path. Defining expanded label set $\mathcal{Y}' = \mathcal{Y} \cup \{\epsilon\}$ where $\epsilon$ is blank symbol, the optimal alignment path $\pi^*$ is obtained via:
\begin{equation}
    \pi^* = \mathop{\arg\max}_{\pi \in \mathcal{Y}'^T} \prod_{t=1}^T P(\pi_t|\mathbf{h}_t) \quad \text{s.t.} \quad \mathcal{B}(\pi^*) = Y
\end{equation}
where $\mathcal{B}$ is the collapsing function that removes blanks and repeats. With such a path, we can obtain token-level alignment between the speech and text. For each token $y_l \in Y$, we can find its temporal boundaries in $\pi^*$:
\begin{equation}
    \begin{aligned}
    \text{A}(y_l) &= \big[ t_{\text{start}}^{(l)}, t_{\text{end}}^{(l)} \big] \\
                      &= \min\{t|\pi^*_t=y_l\}, \max\{t|\pi^*_t=y_l\}.
    \end{aligned}
\end{equation}
Based on this alignment, we organize the response data into a chunk-level speech-text interleaved sequence, with a smaller relative error than word-level interleaving. We adopt a chunk size of 7, which means when segmenting the text, we make cuts at  punctuation marks unless the segment is shorter than 7 words. Appendix \ref{appendix:interleaved} shows an example of such data. It is optional whether to transcribe the speech question into text before generating the response. The total process of constructing the instruction dataset is illustrated in Figure \ref{fig:sft_construction}.

\subsection{Inference}
\label{sec:inference}

The inference process of the CSLM model is illustrated in Figure \ref{fig:architecture}. For an input speech, discrete speech tokens are extracted through the speech tokenizer, and consecutive repeated tokens are merged before being fed into the speech-text joint LLM. The LLM autoregressively outputs a chunk-level text response along with the corresponding speech tokens. The generated speech tokens are then input into the speech decoder, where they are first expanded by the duration predictor and then converted into a Mel-spectrogram by the flow matching model, and finally the audio is synthesized by the HiFi-GAN vocoder. Throughout this process, the LLM continuously outputs interleaved text and speech tokens until completion. Note that there can be a temporal overlap between playing the audio and the model generating the subsequent content \footnote{More details are shown in Appendix \ref{appendix:temporal-overlap}}, which significantly reduces the response latency compared to a full chain-of-modality generation.

\subsection{Language Scalability}
\label{sec:language-scalability}

CSLM models speech using discrete speech tokens. As long as there is parallel speech-text and translation data, a new language can be integrated into CSLM's training. This enables the creation of a speech LLM capable of supporting new languages, indicating that CSLM has excellent scalability in terms of language support.

% It is worth noting that, to ensure robustness to input speech, we use the voices of different speakers (random timbres) to synthesize speech instructions, while we use the voice of a single speaker (a fixed timbre) to synthesize speech responses to ensure consistency.

\section{Experiments}
% 英文表格
\begin{table*}[t]
    \centering
    \resizebox{1.0\linewidth}{!}{
    \begin{tabular}{@{}lccccccccc|c@{}}
        \toprule
        \textbf{Task} & \textbf{Dataset} & \multicolumn{8}{c|}{\textbf{Model}} & \textbf{GT} \\
        \cmidrule(lr){3-10}
        & & Whisper & CosyVoice & SpeechGPT & AnyGPT & GLM-4-Voice & Moshi & \textbf{CSLM-base} & \textbf{CSLM-SFT} & \\
        \midrule
        \textbf{ASR} & \textit{LibriSpeech} & \textbf{2.5} & -- & 18.9 & 8.5 & 2.8 & 5.7 & 6.7 & 9.8 & -- \\
        \textbf{TTS} & \textit{LibriSpeech} & -- & 3.4 & 24.6 & 27.1 & -- & 4.7 & \textbf{3.2} & 3.8 & 3.0 \\
        \textbf{TTS} & \textit{LibriTTS} & -- & \textbf{2.9} & 29.1 & 27.9 & 5.6 & -- & 3.2 & 4.5 & 2.9 \\
        \textbf{TTS} & \textit{VCTK} & -- & 3.7 & 5.7 & 8.5 & -- & -- & 2.8 & \textbf{2.7} & 3.5 \\
        \bottomrule
    \end{tabular}}
    \caption{Results of English ASR and TTS tasks. The test datasets include the \texttt{test-clean} set of LibriSpeech \citep{panaytov2015librispeech}, the \texttt{test-clean} set of LibriTTS \citep{zen19_interspeech}, and VCTK \citep{yamagishi2019cstr}. The Whisper model refers to whisper-large-v3, and the CosyVoice model refers to CosyVoice-300M-SFT. The last column ``GT'' is an abbreviation for ``ground truth'', representing the error rates of the original speech-text pairs from the dataset calculated using whisper-large-v3.}
    \label{tab:base-eval-en}
\end{table*}

\begin{table*}[t]
    \centering
    \small
    \begin{tabular}{@{}lcccccc|c@{}}
        \toprule
        \textbf{Task} & \textbf{Dataset} & \multicolumn{5}{c|}{\textbf{Model}} & \textbf{GT} \\
        \cmidrule(lr){3-7}
        & & Whisper & CosyVoice & GLM-4-Voice & \textbf{CSLM-base} & \textbf{CSLM-SFT} & \\
        \midrule
        \textbf{ASR} & \textit{AISHELL-1} & 9.3 & -- & \textbf{2.5} & 8.6 & 9.0 & -- \\
        \textbf{ASR} & \textit{AISHELL-2} & \textbf{5.4} & -- & -- & 7.6 & 8.6 & -- \\
        \textbf{ASR} & \textit{AISHELL-3} & 14.8 & -- & -- & \textbf{9.2} & 9.6 & -- \\
        \textbf{TTS} & \textit{AISHELL-1} & -- & \textbf{3.3} & -- & 3.8 & 3.7 & 1.9 \\
        \textbf{TTS} & \textit{AISHELL-2} & -- & \textbf{5.0} & -- & 5.3 & 5.2 & 2.8 \\
        \textbf{TTS} & \textit{AISHELL-3} & -- & \textbf{3.8} & -- & 4.9 & 5.3& 2.6 \\
        \bottomrule
    \end{tabular}
    \caption{Results of Chinese ASR and TTS tasks. The test datasets include AISHELL-1 \citep{8384449}, AISHELL-2 \citep{du2018aishell2transformingmandarinasr} and AISHELL-3 \citep{shi21c_interspeech}. The last column ``GT'' represents the error rates of the original speech-text pairs from the dataset calculated using the paraformer large model.}
    \label{tab:base-eval-zh}
\end{table*}

\subsection{Continual Pre-training}
\label{sec:continual-pretraining}

\noindent\textbf{Datasets}~
In the continual pre-training stage, we continue to train the LLM with cross-modal, cross-lingual and mono-lingual text data. All the datasets that we use are open-source ones.
\begin{itemize}
    \item \textbf{Cross-modal Data} We collect parallel Chinese and English speech-text data to form a cross-modal aligned dataset. For English, we use the English subset of Multilingual LibriSpeech \citep{pratap20_interspeech} dataset and the GigaSpeech \citep{chen21o_interspeech} dataset as ASR and TTS data, with half of the examples allocated to each task. For Chinese, we use the WenetSpeech \citep{zhang2022wenetspeech} dataset as the ASR dataset and the WenetSpeech4TTS \citep{ma2024wenetspeech4tts12800hourmandarintts} dataset as the TTS dataset.
    \item \textbf{Cross-lingual Data} For the cross-lingual data, we select a subset from the Chinese-English translation direction in WMT17 \footnote{\url{https://www.statmt.org/wmt17/translation-task.html}}, ensuring that the data count is the same for both translation directions and that the length of each example is medium (see Appendix \ref{appendix:data-preprocessing}).
    \item \textbf{Mono-lingual Instruction Data} For monolingual instruction data, we utilize the InfinityInstruct \footnote{\url{https://github.com/FlagOpen/Infinity-Instruct/tree/main}} dataset, which includes both single-turn and multi-turn instruction data in Chinese and English.
\end{itemize}

The data statistics of CSLM's continual pre-training stage are presented in Table \ref{tab:statistics-pretraining}.
\begin{table}[t]
    \centering
    \resizebox{\linewidth}{!}{
    \begin{tabular}{@{}lccc@{}}
        \toprule
        \textbf{Dataset} & \textbf{Hours} & \textbf{Speech Tokens} & \textbf{Text Tokens} \\
        \midrule
        \multicolumn{4}{c}{\textit{Cross-modal Data}} \\
        \midrule
        MLS English & 44.7K & 3.7B & 0.6B \\
        GigaSpeech & 10.0K & 0.7B & 0.2B \\
        WenetSpeech & 10.0K & 1.1B & 0.5B \\
        WenetSpeech4TTS & 12.8K & 0.6B & 0.2B \\
        \midrule
        \multicolumn{4}{c}{\textit{Cross-lingual Data}} \\
        \midrule
        WMT17 zh-en & - & - & 0.6B \\
         \midrule
        \multicolumn{4}{c}{\textit{Mono-lingual Text Data}} \\
        Infinity-Instruct & - & - & 1.9B \\
        \midrule
        \textbf{Total} & - & \textbf{6.1B} & \textbf{3.9B} \\
        \bottomrule
    \end{tabular}}
    \caption{Statistics of CSLM's continuing pre-training data.}
    \label{tab:statistics-pretraining}
\end{table}

\noindent\textbf{Model Configuration}~
We use Llama-3.1-8B-Instruct \citep{dubey2024llama3} as the foundation LLM. We expand the vocabulary by adding 4,096 speech tokens, matching the vocabulary size of the CosyVoice model.

\subsection{Supervised Fine-tuning (SFT)}
\label{sec:supervised-finetuning}

\noindent\textbf{Datasets}~
We train the model on mono-lingual and cross-lingual speech-to-speech instruction datasets, coupled with mono-lingual text instruction data and cross-lingual translation data for replay.
\begin{itemize}
    \item \textbf{Mono-lingual Speech-to-speech Data} We used the text from the InstructS2S-200K English instruction dataset from \citet{fang2024llamaomniseamlessspeechinteraction}, along with its Chinese translation, as monolingual instruction data. We synthesize these data into speech use CosyVoice-300M-SFT \footnote{\url{https://www.modelscope.cn/models/iic/CosyVoice-300M-SFT}}. More details can be found in the Appendix \ref{appendix:mono-instruct}.
    \item \textbf{Cross-lingual Speech-to-speech Data} We use the Alpaca English instruction dataset and the Chinese translation of Alpaca from \citet{zhu2023extrapolating}, ensuring that each data entry has both English and Chinese versions, allowing us to create cross-lingual instructions. We continue to use CosyVoice-300M-SFT for speech synthesis. Each data entry includes bidirectional English-Chinese instruction/response pairs. The total number of such cross-lingual speech-to-speech data is 104K.
    \item \textbf{Mono-lingual Text Instruction Data} We randomly select a subset of 400K entries from InfinityInstruct used in Section \ref{sec:continual-pretraining}. 
    \item \textbf{Cross-lingual Data} We randomly select a subset of 200K entries from the WMT17 dataset used in \ref{sec:continual-pretraining}.
\end{itemize}

\subsection{Duration Predictor}
\label{sec:duration-predictor}

The duration predictor module in the speech decoder is a two-layer convolutional module that predicts sequence durations of the input speech. 

We train the model using full fine-tuning during both stages. Training details of the LLM and the duration predictor are listed in Appendix \ref{appendix:training-detail}. Details of data preprocessing are in Appendix \ref{appendix:data-preprocessing}.

\section{Evaluation}
\subsection{Basic Tasks}
\label{sec:base-model}

We evaluate CSLM on two basic cross-modal tasks, ASR and TTS. For both tasks, we measure the error rates compared to the ground-truth answers, specifically word error rate (WER) for English and character error rate (CER) for Chinese. We use a speech decoder coupled with a duration predictor to synthesize the generated speech tokens into waveforms, after which the waveforms are transcribed back into text using an ASR model to calculate error rates, resulting in ASR-WER or ASR-CER. For English, we use the Whisper large-v3 \footnote{\url{https://huggingface.co/openai/whisper-large-v3}} \citep{10.5555/3618408.3619590} as the ASR model, while for Chinese we use paraformer large \footnote{\url{https://www.modelscope.cn/models/iic/speech_paraformer-large_asr_nat-zh-cn-16k-common-vocab8404-pytorch}} \citep{gao22b_interspeech}.
% Besides, before computing the error rate, the text predictions or text transcriptions are all normalized.

\begin{table}[t]
    \centering
    \small
    \resizebox{\linewidth}{!}{
    \begin{tabular}{l c}
        \toprule
        \textbf{Model} & \textbf{Speech Data (k hours)} \\
        \midrule
        SpeechGPT* & 11+ \\
        AnyGPT & 57 \\
        Moshi & 7,000+ \\
        GLM-4-Voice** & 10,000+ \\
        \midrule
        CSLM & 77 \\
        \bottomrule
    \end{tabular}}
    \caption{Comparison of amounts of speech data in speech-text pairs used by different models. *The data volume of SpeechGPT is calculated based on the datasets listed in its paper \citep{zhang-etal-2023-speechgpt}. **The data volume of GLM-4-Voice is calculated from the number of speech tokens and its frequency reported in its paper \citep{zeng2024glm4voiceintelligenthumanlikeendtoend}.}
    \label{tab:speech-data}
\end{table}
\begin{table}[t]
    \centering
    \resizebox{\linewidth}{!}{
    \begin{tabular}{lcccccc}
        \toprule
        Model & \multicolumn{2}{c}{GPT Score$\uparrow$} & \multicolumn{1}{c}{UTMOS$\uparrow$} & \multicolumn{1}{c}{ASR-ER$\downarrow$} & \multicolumn{2}{c}{Off-Target$\downarrow$} \\
        \cmidrule(lr){2-3} \cmidrule(lr){6-7}
        & T & S & & & T & S \\
        \midrule
        \multicolumn{7}{c}{En (\textit{InstructS2S-Eval})} \\
        \midrule
        SpeechGPT     & 2.19 & 2.98 & 3.9  & 45.0 & -- & -- \\
        AnyGPT        & -- & 2.31 & 3.3  & --   & -- & -- \\
        GLM-4-Voice   & \textbf{4.10} & \textbf{4.02} & 4.0  & 12.8 & 8.5  & 9.0     \\
        \textbf{CSLM} & 3.50 & 3.27 & \textbf{4.4}  & 9.0  & \textbf{0.0}  & \textbf{0.0} \\
        \midrule
        \multicolumn{7}{c}{Zh (\textit{BELLE-eval-S2S})} \\
        \midrule
        GLM-4-Voice   & \textbf{4.80} & \textbf{4.70} & -- & \textbf{4.6}  & \textbf{0.4}    & 4.8     \\
        \textbf{CSLM} & 3.78 & 3.37 & -- & 6.9  & 0.8    & 5.2 \\
        \bottomrule
    \end{tabular}}
    \caption{Evaluation results of SFT models on English and Chinese speech-to-speech conversational benchmarks. The ``T'' and ``S'' under GPT-Score denote the evaluation of the generated text and the transcription of the generated speech, respectively. The ``T'' and ``S'' under Off-Target denote the off-target ratio assessed for the generated text and speech, respectively. ASR-ER denotes ASR-WER for En or ASR-CER for Zh.}
    \label{tab:sft-en-zh}
\end{table}

\begin{table}[t]
    \centering
    \resizebox{\linewidth}{!}{
    \begin{tabular}{lccccc}
        \toprule
        Model & \multicolumn{2}{c}{GPT Score$\uparrow$} & \multicolumn{1}{c}{ASR-ER$\downarrow$} & \multicolumn{2}{c}{Off-Target$\downarrow$} \\
        \cmidrule(lr){2-3} \cmidrule(lr){5-6}
        & T & S & & T & S \\
        \midrule
        \multicolumn{6}{c}{En$\rightarrow$Zh (\textit{InstructS2S-Eval})} \\
        \midrule
        GLM-4-Voice   & \textbf{4.29} & \textbf{4.23} & \textbf{5.5}   & 8.5  & 8.5     \\
        \textbf{CSLM} & 3.31 & 2.95 & 17.5  & \textbf{1.0}  & \textbf{0.5} \\
        \midrule
        \multicolumn{6}{c}{Zh$\rightarrow$En (\textit{BELLE-eval-S2S})} \\
        \midrule
        GLM-4-Voice   & 1.20 & 1.16 & 42.6  & 97.2   & 96.8    \\
        \textbf{CSLM} & \textbf{3.53} & \textbf{3.20} & \textbf{7.4}   & \textbf{1.6}    &  \textbf{0.8} \\
        \bottomrule
    \end{tabular}}
    \caption{Results of SFT models on En-Zh and Zh-En speech-to-speech conversational tasks.}
    \label{tab:sft-cross}
\end{table}

We compare our CSLM-base and CSLM-SFT model with the base models of SpeechGPT \citep{zhang-etal-2023-speechgpt}, AnyGPT \citep{zhan2024anygptunifiedmultimodalllm}, GLM-4-Voice \citep{zeng2024glm4voiceintelligenthumanlikeendtoend}, and Moshi \citep{défossez2024moshispeechtextfoundationmodel} for English tasks, which are all speech LLMs based on discrete speech tokens. For Chinese, we compare our models with the base model of GLM-4-Voice. We also include results of specialized ASR model whisper-large-v3 and specialized TTS model CosyVoice-300M-SFT.

Results in Table \ref{tab:base-eval-en} and Table \ref{tab:base-eval-zh} show our model outperforms SpeechGPT and AnyGPT, which are fine-tuned with a similar scale of speech-text parallel data. It also achieves a performance comparable to that of Moshi and GLM-4-Voice, both using speech-text pairs dozens or even hundreds of times more than CSLM. The amount of speech data we use is about one percent of that used by GLM-4-Voice and Moshi, as shown in Table \ref{tab:speech-data}. CSLM also performs close to the specialized smaller models. The results indicate that the CSLM base model has strong speech-text alignment capabilities in both English and Chinese.

\subsection{Speech Conversation}
\label{sec:sft-model}

We evaluate CSLM-SFT on mono-lingual and cross-lingual speech-to-speech conversations with both automated metrics and human evaluations.

\subsubsection{Automated Metrics}
For mono-lingual English evaluation, we utilize the \texttt{helpful\_base} and \texttt{vicuna} subsets of AlpacaEval \citep{alpaca_eval}, excluding examples unsuitable for speech interaction, following \citet{fang2024llamaomniseamlessspeechinteraction}. This dataset contains 199 English speech instructions and is referred to as \textit{InstructS2S-Eval}. For mono-lingual Chinese evaluation, we select 250 instructions suitable for speech dialogue scenarios from the BELLE \citep{BELLE} evaluation set, synthesize them into audio using CosyVoice-300M-SFT and create a Chinese speech test set referred to as \textit{BELLE-eval-S2S}. We retain the use of these two sets for cross-lingual evaluation except for instructing the model to respond in the other language. The model's speech-to-speech capability is evaluated from the following three aspects: 

\begin{itemize}
    \item \textbf{Content Quality} We use GPT4o \citep{gpt4o} to score the outputs of the model to evaluate its ability to follow instructions and generate responses. We follow the prompts and setups in \citet{fang2024llamaomniseamlessspeechinteraction}.
    \item \textbf{Speech Quality} To measure the quality of the output speech, we use the UTMOS \citep{saeki22c_interspeech} model to calculate the Mean Opinion Score (MOS), which indicates the naturalness of the English speech. We refer to this metric as the UTMOS score.
    \item \textbf{Speech-Text Consistency} The consistency between speech and text output is measured by calculating error rates, specifically ASR-WER and ASR-CER.
    \item \textbf{Language Accuracy} As a cross-lingual model, CSLM may generate results in unintended languages. We employ the metric of off-target ratio to assess this issue. See Appendix \ref{appendix:lang-detect} for the calculation of this metric.
\end{itemize}

Table \ref{tab:sft-en-zh} presents the results of the mono-lingual speech conversational tasks. CSLM exhibits the best speech naturalness and demonstrates good speech-text consistency along with an extremely low off-target ratio, indicating that CSLM has advantages in cross-modal alignment and language accuracy. The content rating of responses generated by CSLM is better than that of SpeechGPT and AnyGPT. Results of cross-lingual tasks are shown in Table \ref{tab:sft-cross}. In cross-lingual conversations, CSLM still maintains an extremely low off-target ratio. Compared to single-language tasks, there is not much degradation in content quality, demonstrating its cross-language alignment capability.

\begin{table}[t]
    \centering
    \resizebox{\linewidth}{!}{
    \begin{tabular}{lccccc}
        \toprule
        Model & \multicolumn{4}{c}{C-MOS$\uparrow$} & \multicolumn{1}{c}{A-MOS$\uparrow$} \\
        \cmidrule(lr){2-5} \cmidrule(lr){6-6}
        & En$\rightarrow$En & En$\rightarrow$Zh & Zh$\rightarrow$Zh & Zh$\rightarrow$En & Overall \\
        \midrule
        SpeechGPT   & 1.83 & - & - & - & 3.22 \\
        GLM-4-Voice & \textbf{4.17} & \textbf{3.50} & \textbf{4.42} & 3.08 & \textbf{4.11} \\
        \textbf{CSLM} & 3.25 & 3.42 & 4.00 & \textbf{4.33} & 4.00 \\
        \bottomrule
    \end{tabular}}
    \caption{Human evaluation results for C-MOS and A-MOS. En$\rightarrow$X and Zh$\rightarrow$X directions are evaluated on InstructS2S-Eval and BELLE-eval-S2S, respectively.}
    \label{tab:human-eval}
\end{table}

\subsubsection{Human Evaluations}
In addition to automated metrics, we conduct human evaluations to further validate our model's performance on speech-to-speech conversational tasks. We perform a double-blind rating comparing CSLM against baseline models (SpeechGPT and GLM-4-Voice) to assess Content Mean Opinion Scores (C-MOS) and Acoustic Mean Opinion Scores (A-MOS). As shown in Table \ref{tab:human-eval}, CSLM achieves competitive C-MOS across both mono-lingual and cross-lingual pairs. Crucially, the overall trend of these human judgments aligns consistently with our GPT-based and UTMOS evaluations, firmly substantiating the reliability of our automated metrics and demonstrating CSLM's effectiveness in cross-lingual scenarios.

\section{Ablation Study}
\begin{table}[t]
    \centering
    \resizebox{\linewidth}{!}{
    \begin{tabular}{lcc}
        \toprule
        Model & Parallel Speech-Text Pairs & Random Speech-Text Pairs \\
        \midrule
        SpeechGPT   & 1.2\%  & 1.6\%  \\
        GLM-4-Voice & 39.4\% & 16.9\% \\
        \textbf{CSLM} & \textbf{72.5\%} & \textbf{56.7\%} \\
        \bottomrule
    \end{tabular}}
    \caption{Comparison of speech-text representation similarity on LibriSpeech \texttt{test-clean} set.}
    \label{tab:similarity}
\end{table}

\begin{table}[t]
    \centering
    \resizebox{\linewidth}{!}{
    \begin{tabular}{lccccc}
        \toprule
        Model & \multicolumn{2}{c}{GPT Score$\uparrow$} & \multicolumn{1}{c}{ASR-ER$\downarrow$} & \multicolumn{2}{c}{Off-Target$\downarrow$} \\
        \cmidrule(lr){2-3} \cmidrule(lr){5-6}
        & T & S & & T & S \\
        \midrule
        % \multicolumn{6}{c}{En$\rightarrow$En (\textit{InstructS2S-Eval})} \\
        % \midrule
        % CSLM & \textbf{3.50} & \textbf{3.27} & \textbf{9.0}  & \textbf{0.0} & \textbf{0.0} \\
        % \hspace{1em}-- \textit{w/o MT} & 3.21 & 2.96 & 10.4 & \textbf{0.0} & \textbf{0.0} \\
        % \midrule
        % \multicolumn{6}{c}{Zh$\rightarrow$Zh (\textit{BELLE-eval-S2S})} \\
        % \midrule
        % CSLM & \textbf{3.78} & \textbf{3.37} & \textbf{6.9}  & \textbf{0.8} & \textbf{5.2} \\
        % \hspace{1em}-- \textit{w/o MT} & 3.60 & 3.18 & 11.7 & 1.2 & 6.0 \\
        % \midrule
        \multicolumn{6}{c}{En$\rightarrow$Zh (\textit{InstructS2S-Eval})} \\
        \midrule
        CSLM & \textbf{3.31} & \textbf{2.95} & \textbf{17.5} & \textbf{1.0} & \textbf{0.5} \\
        \hspace{1em}-- \textit{w/o MT} & 3.00 & 2.65 & 25.4 & 4.0 & \textbf{0.5} \\
        \midrule
        \multicolumn{6}{c}{Zh$\rightarrow$En (\textit{BELLE-eval-S2S})} \\
        \midrule
        CSLM & \textbf{3.53} & \textbf{3.20} & \textbf{7.4} & 1.6 & 0.8 \\
        \hspace{1em}-- \textit{w/o MT} & 3.08 & 2.74 & 9.2 & \textbf{1.2} & \textbf{0.4} \\
        \bottomrule
    \end{tabular}}
    \caption{Performances of models with and without MT data on speech-to-speech conversational tasks.}
    \label{tab:role-mt}
\end{table}

\subsection{Cross-modal Alignment Efficacy}
\label{sec:cross-modal-alignment}

% To assess cross-modal alignment efficacy during continual pre-training, we measure similarity between speech and text representations of the CLSM-base model on the LibriSpeech \texttt{test-clean} set. Average sentence-level similarity in the last hidden layer for parallel speech-text inputs is 72.5\%, versus 56.7\% for speech and random text, demonstrating effective cross-modal alignment.

To assess cross-modal alignment efficacy during continual pre-training, we compute the speech-text representation similarity for CSLM, SpeechGPT, and GLM-4-Voice on the LibriSpeech \texttt{test-clean} set. We measure the average sentence-level similarity in the last hidden layer for both parallel speech-text pairs and random speech-text pairs. As shown in Table \ref{tab:similarity}, CSLM achieves a superior parallel speech-text similarity of 72.5\%, compared to GLM-4-Voice (39.4\%) and SpeechGPT (1.2\%). For speech and random text pairs, CSLM scores 56.7\%. This high baseline similarity for random pairs likely stems from the shared representation space developed during continual pre-training, where dense, language- and modality-agnostic embeddings inherently align text and speech. However, the substantial gap between CSLM's parallel (72.5\%) and random (56.7\%) scores, combined with its superiority over baseline models, affirms the specific, fine-grained cross-modal alignment established by our training methodology.

\subsection{Effect of MT Data}
\label{sec:effect-mt-data}

To measure the effect of MT data during the continual pre-training stage, we conduct an ablation experiment by training a model without MT data during the continual pre-training stage, followed by same instruction fine-tuning process with CSLM. It can be observed from Table \ref{tab:role-mt} that the model trained without MT data exhibits lower content quality in cross-lingual tasks. Additionally, it performs worse in terms of ASR-ER, indicating a decline in the quality of the generated speech content.

\begin{table}[t]
    \centering
    \resizebox{\linewidth}{!}{
    \begin{tabular}{lccccc}
        \toprule
        Model & \multicolumn{2}{c}{GPT Score$\uparrow$} & \multicolumn{1}{c}{ASR-ER$\downarrow$} & Latency(s)$\downarrow$ & Speedup$\uparrow$ \\
        \cmidrule(lr){2-3}
        & T & S & & & \\
        \midrule
        \multicolumn{6}{c}{En$\rightarrow$En (\textit{InstructS2S-Eval})} \\
        \midrule
        CSLM & \textbf{3.50} & \textbf{3.27} & 9.0 & 466.46 & $\times$2.87 \\
        \hspace{1em}-- \textit{chunk=4} & 2.82 & 2.42 & 15.2 & \textbf{456.88} & \textbf{$\times$2.93} \\
        \hspace{1em}-- \textit{full CoM} & 3.21 & 2.92 & \textbf{8.5} & 1338.68 & $\times$1 \\
        \hspace{1em}-- \textit{w/o TQ} & 2.01 & 1.20 & 8.9 & -- & -- \\
        \midrule
        \multicolumn{6}{c}{Zh$\rightarrow$Zh (\textit{BELLE-eval-S2S})} \\
        \midrule
        CSLM & \textbf{3.78} & 3.37 & 6.9 & 631.76 & $\times$1.92 \\
        \hspace{1em}-- \textit{chunk=4} & 2.66 & 2.29 & 23.5 & \textbf{620.17} & \textbf{$\times$1.96} \\
        \hspace{1em}-- \textit{full CoM} & 3.68 & \textbf{3.42} & \textbf{6.1} & 1215.54 & $\times$1 \\
        \hspace{1em}-- \textit{w/o TQ} & 2.12 & 2.00 & 6.6 & -- & -- \\
        \midrule
        \multicolumn{6}{c}{En$\rightarrow$Zh (\textit{InstructS2S-Eval})} \\
        \midrule
        CSLM & \textbf{3.31} & \textbf{2.95} & 17.5 & 437.88 & $\times$4.62 \\
        \hspace{1em}-- \textit{chunk=4} & 1.62 & 1.55 & 34.7 & \textbf{435.32} & \textbf{$\times$4.65} \\
        \hspace{1em}-- \textit{full CoM} & 3.27 & 2.92 & 11.8 & 2024.48 & $\times$1 \\
        \hspace{1em}-- \textit{w/o TQ} & 1.25 & 1.23 & \textbf{10.9} & -- & -- \\
        \midrule
        \multicolumn{6}{c}{Zh$\rightarrow$En (\textit{BELLE-eval-S2S})} \\
        \midrule
        CSLM & \textbf{3.53} & \textbf{3.20} & \textbf{7.4} & 666.40 & $\times$2.30 \\
        \hspace{1em}-- \textit{chunk=4} & 2.74 & 2.45 & 32.1 & \textbf{601.46} & \textbf{$\times$2.55} \\
        \hspace{1em}-- \textit{full CoM} & 3.05 & 2.79 & 10.7 & 1531.16 & $\times$1 \\
        \hspace{1em}-- \textit{w/o TQ} & 1.22 & 1.19 & 7.8 & -- & -- \\
        \bottomrule
    \end{tabular}}
    \caption{Impact of chain-of-modality forms on speech-to-speech conversations.}
    \label{tab:form-com}
\end{table}

\subsection{Form of Chain-of-modality}
\label{sec:form-com}

We compare chain-of-modality generation processes containing different components to validate the effectiveness of our speech-text interleaved chain-of-modality. We train a model with a chunk size of 4 in the interleaved chain-of-modality to test whether alignment accuracy would reduce performance, as alignment errors can occur at both the beginning and end of each chunk. In addition, we train a model with full CoM (i.e., generating the complete text answer and then generating the complete speech answer) during SFT to validate the performance improvement and speedup effect of our interleaving generation approach. We also train a model that skips generating the text question. Results in Table \ref{tab:form-com} show that: (i) Model with chunk size of 4 performs poorly, indicating that a low-accuracy alignment would severely damage performance. (ii) CSLM with speech-text interleaved chain-of-modality generally outperforms the full CoM model, and can bring about an average speedup of $\times$ 2.93, which validates the efficacy of this interleaved chain-of-modality. CSLM's slight advantage over full CoM model stems from its alignment granularity, which better matches the continual pre-training stage in terms of data length, as full chain-of-modality data can be very long. (iii) CSLM with no text questions shows a significant decline in metrics across all tasks, especially in cross-lingual ones, indicating that cross-language alignment occurs in the text modality, which is in accord with our alignment strategy.

\section{Conclusion}
We introduce CSLM, a cross-lingual speech language model. CSLM comprises a speech tokenizer, speech-text joint LLM and speech decoder, trained with an efficient method featuring cross-modal and cross-lingual alignment. Through continual pre-training and instruction fine-tuning with speech-text interleaved chain-of-modality, CSLM achieves strong cross-modal and cross-lingual alignment, enabling mono- and cross-lingual speech conversation and expanding speech LLMs' applications.

\section*{Limitations}
Due to limitations in available data resources and computational resources, CSLM has not been trained on larger-scale speech and text datasets, leaving its full potential temporarily unverified. Additionally, as a cross-lingual speech model, CSLM still requires expansion to support more languages to further broaden its application scope.

\section*{Acknowledgments}
We thank all the anonymous reviewers for their valuable comments on this paper. This work is supported by the grant from the Beijing Natural Science Foundation (No. L257006).

\bibliography{custom}

\appendix

\section{Interleaved Data Example}
\label{appendix:interleaved}
\begin{tcolorbox}
    [title=Speech-Text Interleaved Data Example ,fontupper=\ttfamily,colback=gray!10!white,colframe=black,arc=1mm,boxrule=1pt,left=1mm,right=1mm,top=1mm,bottom=1mm, fonttitle=\small]
    \small
    \textbf{Prompt}: Please directly answer the questions in the user's speech. This is input: <sosp><1490>...<1947><eosp>.
    
    \textbf{Response}: [question]: Hey, can you think of a, like, really creative way to use just one single pencil?; [answer]: Use it as a plant marker in your garden,<sosp><1555>...<4450><eosp>write the name of each plant on the pencil and stick it in the soil next to it.<sosp><2937>...<4431><eosp>

    {\footnotesize \color{gray!60!black} \quad \\*[3pt] \textit{Note: Content within ellipses (...) represents speech tokens.}}
    
\end{tcolorbox}

% \section{Statistics of Continual Pre-training Data}
% \label{appendix:statistics-pretraining}
% The data statistics of CSLM's continual pre-training stage are presented in Table \ref{tab:statistics-pretraining}.
% \input{latex/table/statistics_pretraining}

\section{Temporal Overlap}
\label{appendix:temporal-overlap}
We provide a concrete example of how much ``temporal overlap'' occurs between playing the generated audio and producing subsequent content. The question is ``How do I wrap a present neatly?'', and the generated answer of CSLM is:
\begin{tcolorbox}
    [title=Example of Temporal Overlap ,fontupper=\ttfamily,colback=gray!10!white,colframe=black,arc=1mm,boxrule=1pt,left=1mm,right=1mm,top=1mm,bottom=1mm, fonttitle=\small]
    \small
    \textbf{[question]: How do I wrap a present neatly?; [answer]: To neatly wrap a present,<sosp><68><1868><1342><2323><773><2621> <2554><3489><940><16><1136><2796><717><1454> <760><822><2537><351><824><2110><2113><870> <2110><690><822><3274><2999><2409><3887><492> <2876><1688><60><302><624><eosp>} start by wrapping the paper or tissue paper around the item,<sosp><1577><1446><3898><1117><124><646> <3049><4><4064><3614><1122><2392><1949><51> <1343><202><266><2293><489><760><822><345> <740><2307><3229><409><2162><2103><101><3684> <1915><1406><1698><2583><942><1122><39><1> <2735><2809><2148><760><1716><1712><477><701> <1618><3740><1507><39><1915><3014><1353><489> <700><758><760><2515><3211><1784><870><822> <3274><1404><193><800><3278><2796><1038><535> <714><1404><109><33><4064><3582><3347><2230> <162><eosp>...
\end{tcolorbox}
Once the text question, the first text response sequence and the first speech response sequence, i.e., the bolded parts, are generated, the already-produced speech tokens can be used to synthesize the speech waveform, and the corresponding audio is then played. Meanwhile, CSLM continues generating the non-bolded portion, resulting in the temporal overlap.

% \section{Speech Data Amount}
% \label{appendix:speech-data}
% Table \ref{tab:speech-data} shows the amounts of speech data in speech-text pairs used by different models for cross-modal alignment. The amount of speech data we use is about one percent of that used by GLM-4-Voice and Moshi.
% \input{latex/table/speech_data}

\section{Mono-lingual Instruction Data}
\label{appendix:mono-instruct}
For English, we adopt the text data of the InstructS2S-200K from \citet{fang2024llamaomniseamlessspeechinteraction}. This dataset encompasses approximately 200K instruction data entries sourced from the Alpaca \citep{taori2023stanford} and UltraChat \citep{ding-etal-2023-enhancing} datasets, with instructions rewritten by an LLM and responses generated by an LLM as well. For Chinese, we utilize the Qwen2.5-72B-Instruct \citep{qwen2025qwen25technicalreport} model to translate the InstructS2S-200K into Chinese, thereby creating a Chinese instruction dataset. Finally, we use CosyVoice-300M-SFT to synthesize speech instructions and responses. For the instructions, we use random timbres generated by the fish-speech 1.5 \footnote{\url{https://huggingface.co/fishaudio/fish-speech-1.5}} \citep{liao2024fishspeechleveraginglargelanguage} model, while for the responses we employ a fixed timbre to ensure consistency.

\section{Training Details}
\label{appendix:training-detail}
At the continual pre-training stage, we train the model with a batch size of 288 for 1 epoch. We use a cosine learning rate scheduler, where the maximum learning rate is set to 6e-5 with the first 3\% of the training steps for warm-up. The maximum sequence length of the model is 2,048. At the supervised fine-tuning stage, we train the model with a batch size of 48 for 1 epoch, and we set the maximum sequence length of the model to 4,096 and the maximum learning rate to 1e-5. The other training setups remain the same as in the first stage. All of training tasks above are conducted using DeepSpeed \footnote{\url{https://github.com/deepspeedai/DeepSpeed}} ZeRO Stage 1 on 24 NVIDIA H800 80G GPUs.

When training the duration predictor module, we use the English speech dataset LJSpeech-1.1 \footnote{\url{https://keithito.com/LJ-Speech-Dataset/}} and the Chinese speech dataset Chinese Standard Mandarin Speech Corpus \footnote{\url{https://www.data-baker.com/open_source.html}} from Baker, which contain 13,100 and 10,000 data entries respectively. We train the module on these datasets for 15 epochs.

\section{Data Preprocessing}
\label{appendix:data-preprocessing}
For machine translation data in the continual pre-training stage, we filter out data where the sum of the source and target lengths is less than 128 or greater than 2048, ensuring that each example's length is medium.

We use the CosyVoice-300M-25hz \footnote{\url{https://www.modelscope.cn/models/iic/CosyVoice-300M-25Hz}} model as the speech tokenizer, which extracts discrete speech tokens from the waveform at a frequency of 25Hz. For the extracted speech tokens, we merge the consecutive repeated ones to improve training efficiency.

In the continual pre-training stage, each example is formatted as an instruction. Following \citet{zhang-etal-2023-speechgpt}, we employ GPT-4o \citep{gpt4o} to generate ASR, TTS, and MT instructions, with a total of 10 instructions for each task. Some of these instructions are as follows.
\begin{tcolorbox}
    [title=ASR \& TTS \& MT Instructions ,fontupper=\ttfamily,colback=gray!10!white,colframe=black,arc=1mm,boxrule=1pt,left=1mm,right=1mm,top=1mm,bottom=1mm, fonttitle=\small]
    \small
    \textbf{ASR (en)}: \\
    Convert the following audio into written English text. \\
    Decode the English phrases from the attached audio. \\
    ... \\
    \textbf{TTS (en)}: \\
    Convert this English text into speech. \\
    Produce English audio from the given text. \\
    ... \\
    \textbf{MT (en $\rightarrow$ zh)}: \\
    Convert the English text below into Chinese. \\
    Change the English content below into Chinese. \\
    ...
    
\end{tcolorbox}

When constructing speech-text interleaved data for supervised fine-tuning, we employ pre-trained speech encoder to get the alignment. For English data, we utilize the wav2vec 2.0 \footnote{Here we refer to WAV2VEC2\_ASR\_BASE\_960H (\url{https://docs.pytorch.org/audio/stable/generated/torchaudio.pipelines.WAV2VEC2_ASR_BASE_960H.html}).} \citep{NEURIPS2020_92d1e1eb} model, while for Chinese data, we use the SenseVoice Small \citep{an2024funaudiollmvoiceunderstandinggeneration} model as the CTC aligner.

\section{Calculation of Off-target Ratio}
\label{appendix:lang-detect}
The specific process to get off-target ratio involves employing an external language detection tool to identify the languages present in the model's generated responses and calculating the ratio of samples that do not match the intended language. We utilize various external tools to detect the language of text responses and speech responses. For the text part of the response, we use \textit{langid} \footnote{\url{https://github.com/saffsd/langid.py}} to detect the language. For the speech part, we use the SenseVoiceSmall \citep{an2024funaudiollmvoiceunderstandinggeneration} model.

% This is an appendix.

\end{document}